\documentclass[tinyml]{acmart}

\usepackage{caption}
\usepackage{subcaption}
\usepackage{array}
\usepackage{ragged2e}

\AtBeginDocument{%
  \providecommand\BibTeX{{%
    \normalfont B\kern-0.5em{\scshape i\kern-0.25em b}\kern-0.8em\TeX}}}

\setcopyright{rightsretained}
\copyrightyear{2022}
\acmYear{2022}

\begin{document}

% \title{Ultra Low Energy Image Segmentation using Parallel CNN Processors}
% \title{L\textsuperscript{3}U-net: Low-Latency Lightweight U-net for Parallel CNN Processors}
\title{L\textsuperscript{3}U-net: Low-Latency Lightweight U-net Based Image Segmentation Model for Parallel CNN Processors}

%
% The "author" command and its associated commands are used to define
% the authors and their affiliations.
% Of note is the shared affiliation of the first two authors, and the
% "authornote" and "authornotemark" commands
% used to denote shared contribution to the research.
\author{Osman Erman Okman}
\authornote{Both authors contributed equally to this research.}
\email{erman.okman@analog.com}
\author{Mehmet Gorkem Ulkar}
\authornotemark[1]
\email{gorkem.ulkar@analog.com}
\author{Gulnur Selda Uyanik}
\email{selda.uyanik@analog.com}
\affiliation{%
  \institution{Analog Devices Inc.}
  \streetaddress{Eski Büyükdere Cd. No:1}
  \city{Istanbul}
  \state{}
  \country{Turkey}
  \postcode{34398}
}

% \author{Anonymous}

\renewcommand{\shortauthors}{Okman and Ulkar, et al.}
\newcommand{\ournet}{L\textsuperscript{3}U-net }

\begin{abstract}
  In this research, we propose a tiny image segmentation model, L\textsuperscript{3}U-net, that works on low-resource edge devices in real-time. We introduce a data folding technique that reduces inference latency by leveraging the parallel convolutional layer processing capability of the CNN accelerators. We also deploy the proposed model to such a device, MAX78000, and the results show that \ournet achieves more than 90\% accuracy over two different segmentation datasets with 10 fps.% when deployed on an edge CNN accelerator.

\end{abstract}

%%
%% Keywords. The author(s) should pick words that accurately describe*

\maketitle

\section{Introduction}
Semantic segmentation is one of the fundamental computer vision tasks that assign each pixel of an image with a predefined label. This task is effectively utilized for many applications like video surveillance, autonomous vehicle guidance, robotics, and biomedical image analysis. For most of these applications, real-time processing with high-resolution inputs and the low energy consumption is critical as many of these applications are designed to be a part of battery-powered devices like drones or personal gadgets. This makes TinyML research essential to develop applications that can run on hardware using mWs of power, KBs of RAM, and less than KBs of flash \cite{tinyml_review}.

With the advance of deep learning (DL), the performance of semantic segmentation models has been improved significantly and now very accurate results are possible. There are various approaches in the literature for this task that differ in particular aspects such as network architecture, cost function, training strategies, training data, etc. \cite{segment_survey}. The initial approaches adopt fully convolutional networks (FCN) \cite{Long_2015_CVPR, Liu_ParseNet} where the global context information cannot be efficiently used. To add more contextual information to FCN approaches, Conditional Random Fields (CRFs) and Markov Random Fields (MRFs) are integrated back to the DL approaches. Liu et al. proposed a CNN model that enables MRFs in a single forward pass \cite{Liu_2015_ICCV}.

One of the most popular image segmentation approaches uses convolutional encoder-decoder model architectures. Noh et al. \cite{Noh_2015_ICCV} proposed to use transposed convolution (deconvolution) layers to reconstruct the segmentation map from the VGG 16-layer network and present promising results on PASCAL VOC 2012 dataset. U-net \cite{unet} and V-Net \cite{vnet} are popular auto encode-based approaches. These architectures are different from \cite{Noh_2015_ICCV} as there are residual connections between specified encoding and decoding layers. Various modified U-net models have also been used for many different applications like 3D image segmentation \cite{3d_unet} or land use determination \cite{land_use_munet, unet_sea_land_seg, ulmas2020segmentation}. U-net is still a prevalent approach, and many recent studies propose modifications like adding dense skip connections \cite{unet3+} or proposing attention-based U-net architectures \cite{sa_unet, attention_unet} for different applications. Weng et al. \cite{nas_unet} proposed a neural architecture search (NAS) approach to search highly accurate U-net-based models for medical image segmentation effectively.

DeepLab \cite{deeplab} family models are another group of commonly preferred semantic segmentation DL models that effectively utilize dilated (a.k.a. atrous) convolutions. This operator controls the resolution at which the features are computed in the DL models. By increasing the dilation parameter, the field of the view of the utilized kernels is increased to deal with larger context effectively without increasing the number of model parameters. Further improvements are proposed to this model \cite{deeplab_v2, deeplab_v3, deeplab_v3+}. Finally, DeepLab v3+, which combines cascaded and parallel dilated convolution modules and uses an auto-encoder approach, achieved an 89\% mean intersection over union (mIoU) score.

A study \cite{segment_survey} compares many other approaches like attention-based, generative or recurrent neural networks, as well as other popular approaches like RefineNet \cite{refine_net}. Nevertheless, these methods are not deployable to low-power, low-memory edge devices due to their high architectural and computational requirements. Most of them cannot achieve real-time inference speeds. Besides, the existing studies on TinyML semantic segmentation are minimal. \cite{esp_net} proposes to use a new convolutional module, efficient spatial pyramid (ESP), efficient computation, memory, and power. A human-machine collaborative design strategy is used in \cite{edgesegnet} to create networks with customized module-level macro architecture and micro-architecture designs specifically for semantic segmentation tasks. A modified dilated convolution approach is also proposed in \cite{liteseg}, where many layers are optimized for this task, and the backbone is changed with MobileNetv2 \cite{mobilnet_v2}. Despite these approaches reducing the computational load and testing on edge, they are not applied on TinyML hardware. Bahl et al. proposed a lightweight encoder-decoder model with depthwise separable convolutions and decreasing memory usage by removing skip connections between encoder and decoder \cite{Bahl_2019_ICCV}. The model is also realized on an FPGA, and its results are given on very low-resolution satellite images. Another recent work demonstrates an attention condenser neural network, AttendSeg, and its results on CamVid dataset \cite{wen2021attendseg}, which is stated as deployable to a tiny edge device.

This study introduces a very lightweight U-net-based encoder-decoder model to solve the semantic segmentation problems on tiny battery-powered hardware. A data reforming (i.e., folding) approach is also proposed to be used for hardware composed of parallel CNN processor units. This strategy also enables the hardware to handle higher spatial resolution images. The complete design is implemented on a low power edge device, MAX78000 \cite{max78000_datasheet}, which is shown as the most energy-efficient CNN inference hardware in the market \cite{battery_powered_face_rec, tiny_ml_hardware_bench, max78000_nas_kws}. To summarize, our contributions are:
\begin{itemize}
\item A fully CNN very lightweight modified U-net model for semantic segmentation problems,
\item A data folding approach that enables higher resolution throughput for parallel CNN processing hardware architectures,
\item An implementation of the proposed approach on such a device, MAX78000, and presentation of the results in terms of accuracy, speed, and energy consumption.
\end{itemize}

The remainder of this paper is organized as follows: In Section \ref{sec:data_folding}, we present the proposed data folding technique. The proposed network architecture is detailed in Section \ref{sec:proposed_arch}. The model deployment platform is described in Section \ref{sec:platform}.
We evaluate the proposed models in Section \ref{subsec:results} trained with the datasets presented in Section \ref{subsec:dataset}. The training approach and parameters are presented in Section \ref{subsec:training}. Finally, Section \ref{sec:conclusion} summarizes the results.

\section{Data Folding Technique}
\label{sec:data_folding}
Inference latency plays a critical role for neural network accelerators at the edge since it directly affects those resource-constrained devices' frame-per-second (FPS) performance metrics. The convolution operation is parallelizable in the channel domain, and this feature has already been utilized in neural network accelerators \cite{max78000_datasheet}. The initial layers of CNN networks generally have the highest spatial resolution and spatial size, whereas they have the lowest channel counts compared to deeper layers of the network. These initial layers induce a significant proportion of the entire network's latency since only a small number of cores can be used to process the small number of input channels that are big in spatial size. However, the spatial resolutions decrease for most network architectures, and the channel counts increase as the network proceeds over the following layers. Therefore, these middle layers are more suitable for parallel processing. U-net is an example of such networks, and the highest spatial resolution is in the initial layers. In the original U-net architecture, the input image consists of only one channel, and its size is 572x572 \cite{unet}. Suppose U-net is deployed in a neural network accelerator. In that case, only one processor core out of many will be active and busy processing the big input, which generates a significant amount of all network latency.

\begin{figure}[h]
  \centering
  \includegraphics[width=\linewidth]{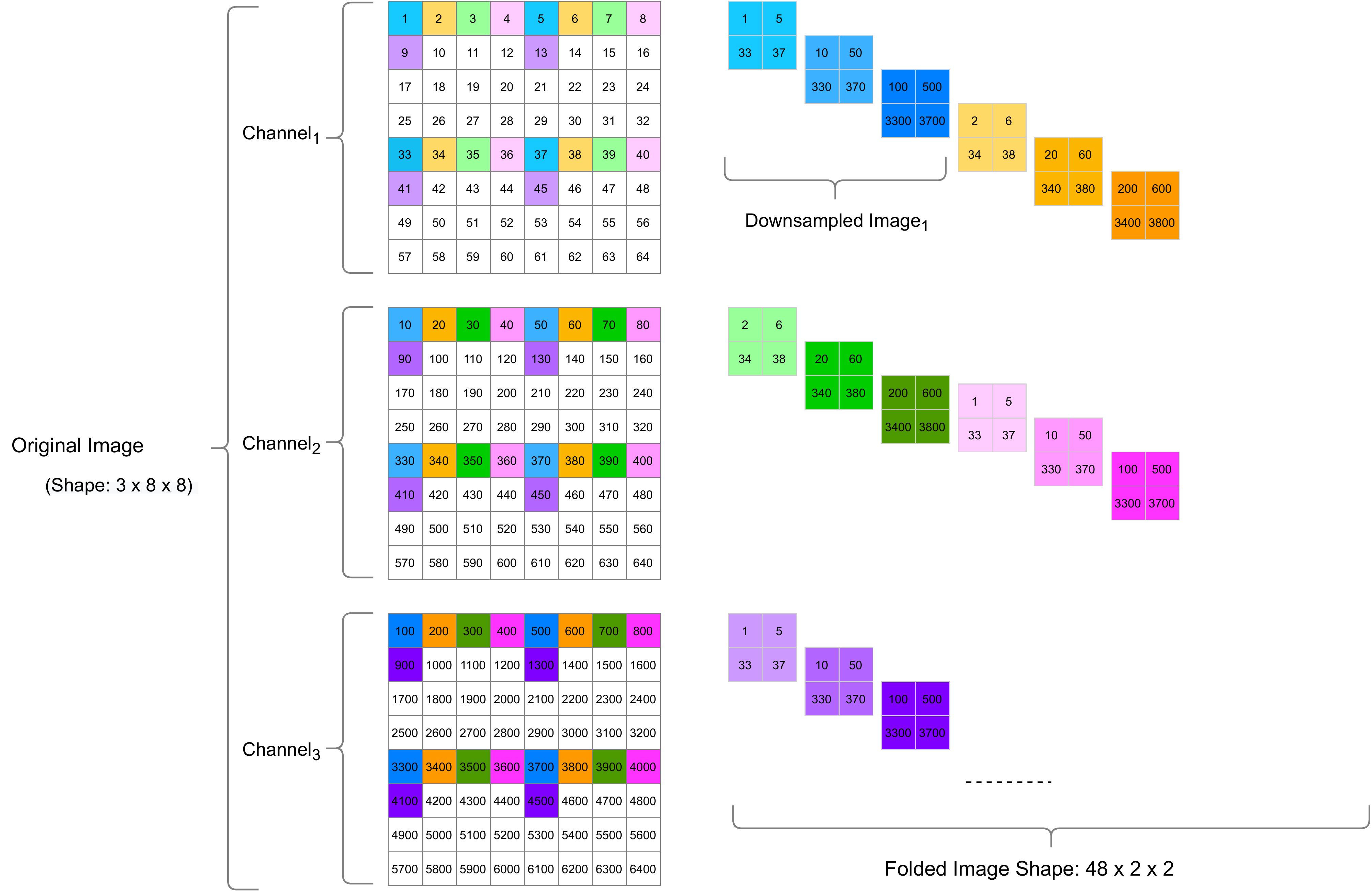}
  \caption{Visualization of the proposed data folding technique}
  \label{fig:folding}
\end{figure}

\begin{figure}[h]
  \centering
  \includegraphics[width=\linewidth]{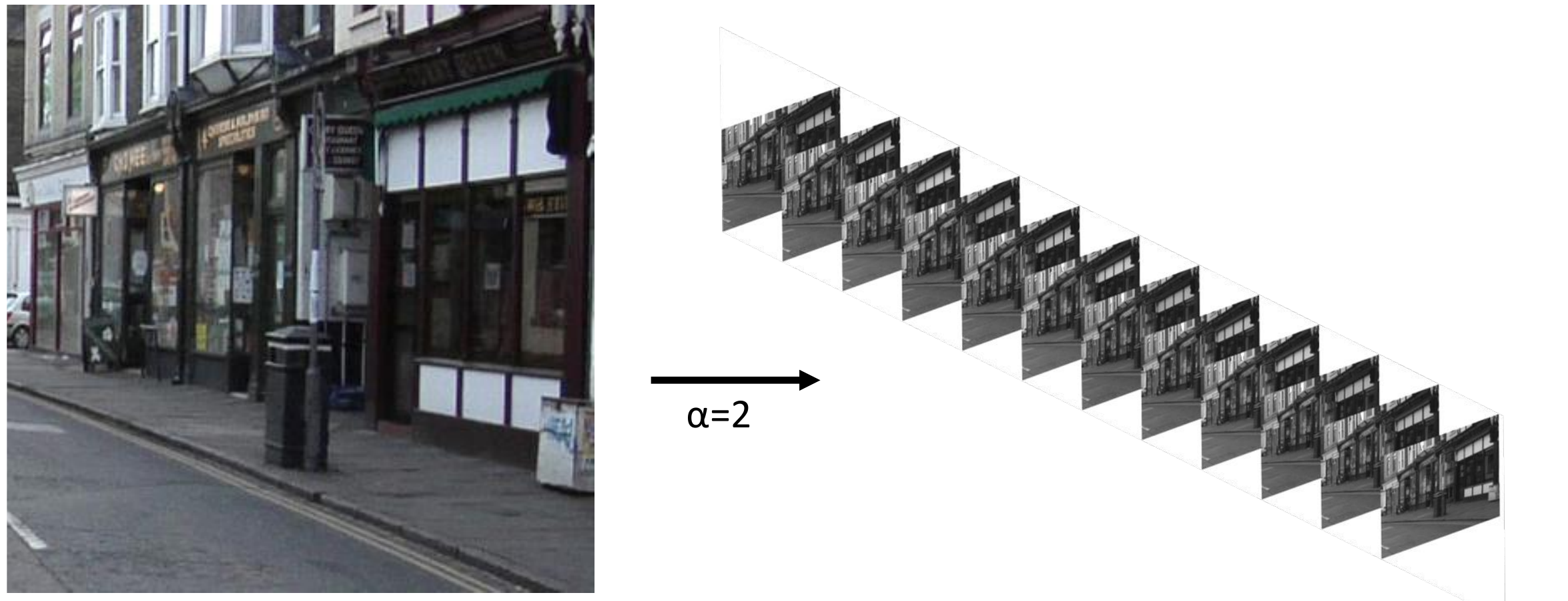}
  \caption{Example for input image folding for $\alpha$=2. The original image is 3x352x352 and the folded image is 12x176x176.}
  \label{fig:folding_example}
\end{figure}

To optimize the latency of the initial layers and to distribute the uneven processor loads to many processor cores more evenly, we propose a novel technique, data folding. Applying this technique and performing a regular convolution with stride=1 is mathematically equivalent to strided convolutions; however, the latency is reduced by distributing the processing load to the parallel cores. The proposed data folding technique works by creating downsampled versions of the input data by first shifting in height and width, then combining those downsampled versions of the original data in the channel domain. Although the spatial resolution appears to shrink, all of the original information is fed to the network by increasing the channel count. Consequently, the data becomes more suitable for parallel processing since more channels mean more processing cores can be utilized for the same amount of data processing. Likewise, the processing latency drops as each processor needs to handle smaller data due to lowered resolution after folding. In Fig. \ref{fig:folding}, the folding of input data of 3x8x8 size into 48x2x2 is illustrated, where the shapes are given as channel size x height x width format (CHW). As shown, the channels of the folded data are the downsampled versions of the original channels. In this example, the folding factor, $\alpha$, is assumed to be 4. The downsampled data from each channel with sampling offsets are placed through the channels of the new kernel, where the sampling offsets vary between 0 to $\alpha-1$ in both height and width axes. In Fig \ref{fig:folding}, blue channels show the downsampled data with sampling offset is equal to 0 for both height and width. On the other hand, yellow channels illustrate the downsampled data with sampling offset is equal to 0 for the height but 1 for the width. Yellow channels follow the blue channels stacked in the channel axis. Likewise, other downsampled versions with increasing offset values are stacked further. Figure \ref{fig:folding_example} shows an example of a folded image when $\alpha$ is 2.

\begin{figure}[h]
  \centering
  \includegraphics[width=\linewidth]{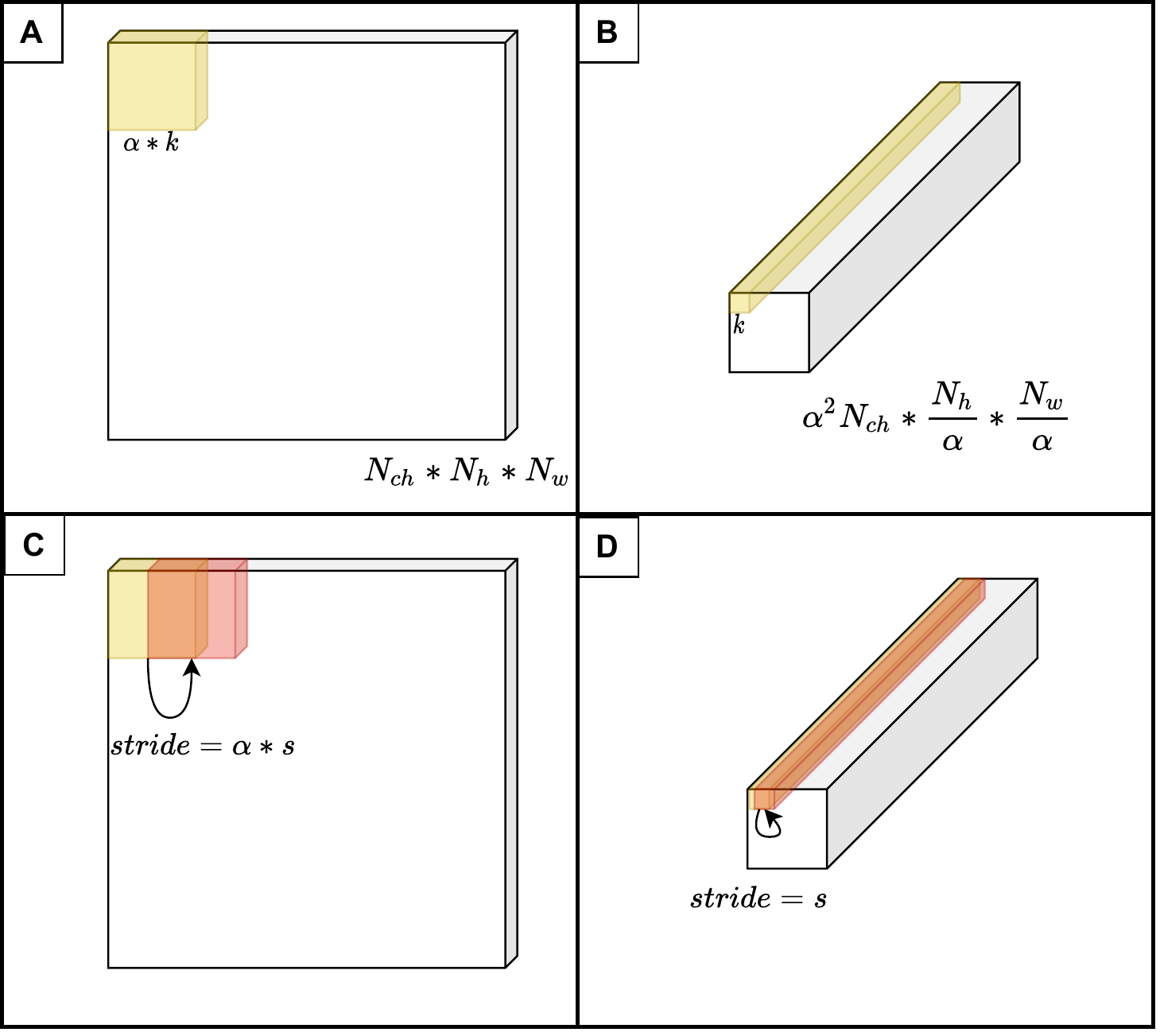}
  \caption{2D convolution operation before and after folding}
  \Description{Data folding }
  \label{fig:folding_4figs}
\end{figure}

\begin{figure}[h]
  \centering
  \includegraphics[width=0.8\linewidth]{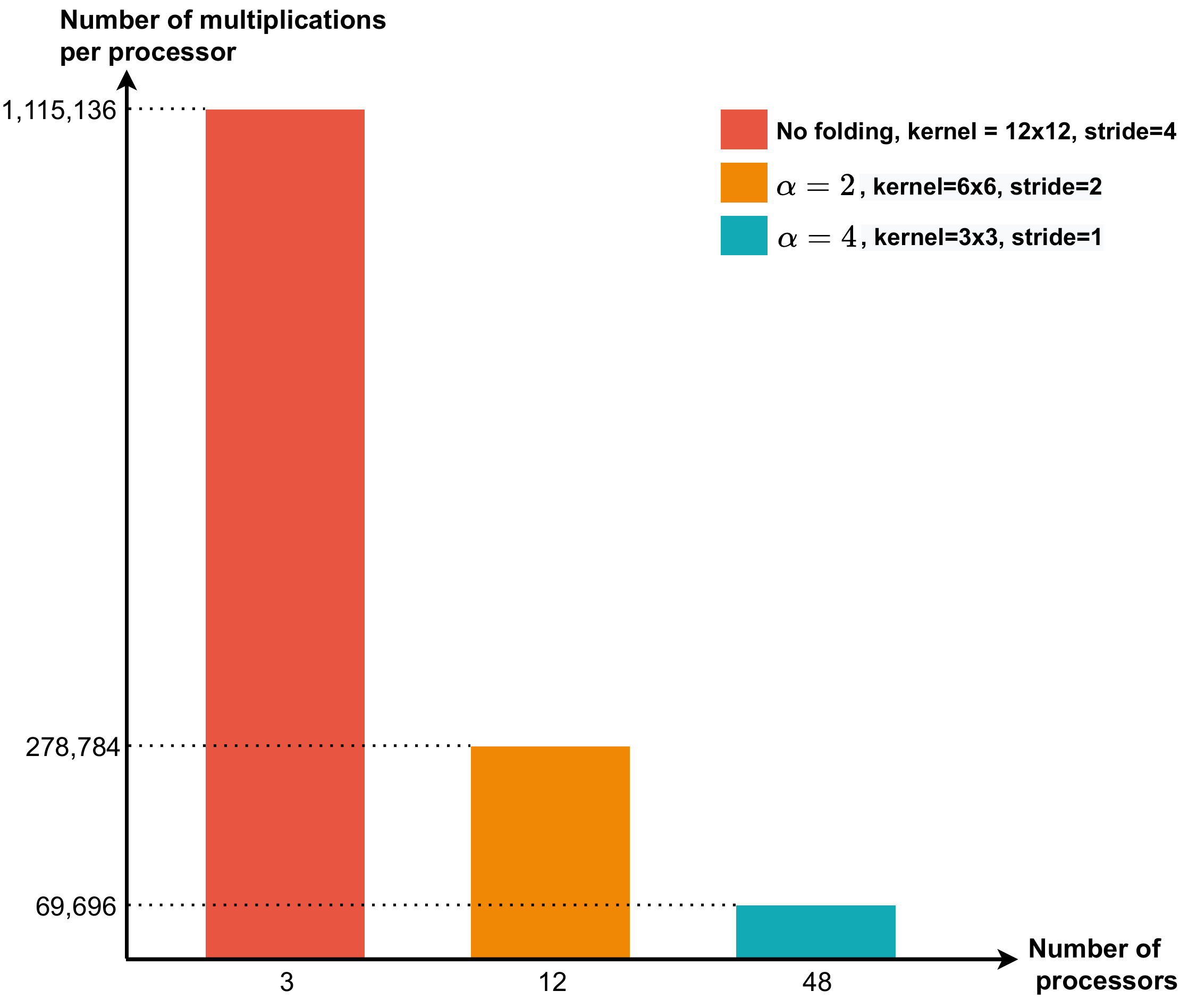}
  \caption{Distribution of convolution operations per processor with regard to the number of input data channels.}
  \label{fig:folding_barchart}
\end{figure}

Since the data in the neighbor pixels are folded into channels, the same data portion can be processed with a smaller 2d convolutional kernel after the folding. In Fig. \ref{fig:folding_4figs} A, the data with a shape $N_{ch}*N_{h}*N_{w}$ is shown. The 2d convolutional kernel with a side length equal to $\alpha*k$ is illustrated with a yellow square prism in the same figure. After folding, the 3d data matrix takes the shape of $\alpha^2N_{ch}*\frac{N_{h}}{\alpha}*\frac{N_{w}}{\alpha}$. The data in the slice of the kernel highlighted in yellow in Fig. \ref{fig:folding_4figs} A is now placed in another square prism. The side length of this prism is now $k$ instead of $\alpha*k$ as seen in Fig. \ref{fig:folding_4figs} B. Therefore, the same data can be accessed with a smaller convolutional kernel after folding. Data folding not only changes the kernel lengths, but also the stride parameter. Since the spatial dimensions shrink after folding, any step in these shrunk dimensions corresponds to larger steps in the original data tensor. Numerically, a convolution stride of $s$ in the folded tensor corresponds to a stride of $\alpha*s$ in the original unfolded tensor as seen in Fig. \ref{fig:folding_4figs} D and C. Thus, data folding is a data reshaping operation that when a 2d convolution operation follows it, it becomes equivalent to a convolution operation on the original tensor with a larger kernel and a larger stride.
As mentioned above, data folding allows distributing the convolution processing load to a larger number of parallel processors. In Fig. \ref{fig:folding_barchart}, the number of multiplications done by each parallel processor for a 2d convolution operation on 352x352 data is shown. The red bar shows the case where there is no data folding. The yellow bar shows the case with folding and the folding factor, $\alpha$, is equal to 2. Lastly, the turquoise bar shows the case for $\alpha=4$. The input data is assumed to have three channels, e.g., RGB channels of an image, and the processor is assumed to have at least 48 parallel cores. It is noteworthy that as $\alpha$ doubles, the processor load on each active processor is quartered if enough idle processors exist. The folding factor must be chosen based on the number of parallel cores and the desired stride on the original tensor. Higher folding factors results in greater steps in the shift of the convolutional kernel in the original tensor.

\section{Proposed \ournet Architecture}
\label{sec:proposed_arch}
The U-net model \cite{unet} is a CNN model that has a condensing path to obtain the context and a symmetrical expanding path to provide accurate localization. There are connections between these paths at different resolutions in the U-net model to allow information shared through the network. To achieve low latency and energy while providing high performance in edge devices, we propose a model that combines the data folding technique with a small version of the U-net architecture.

\begin{figure*}[h]
  \centering \includegraphics[width=1.0\linewidth]{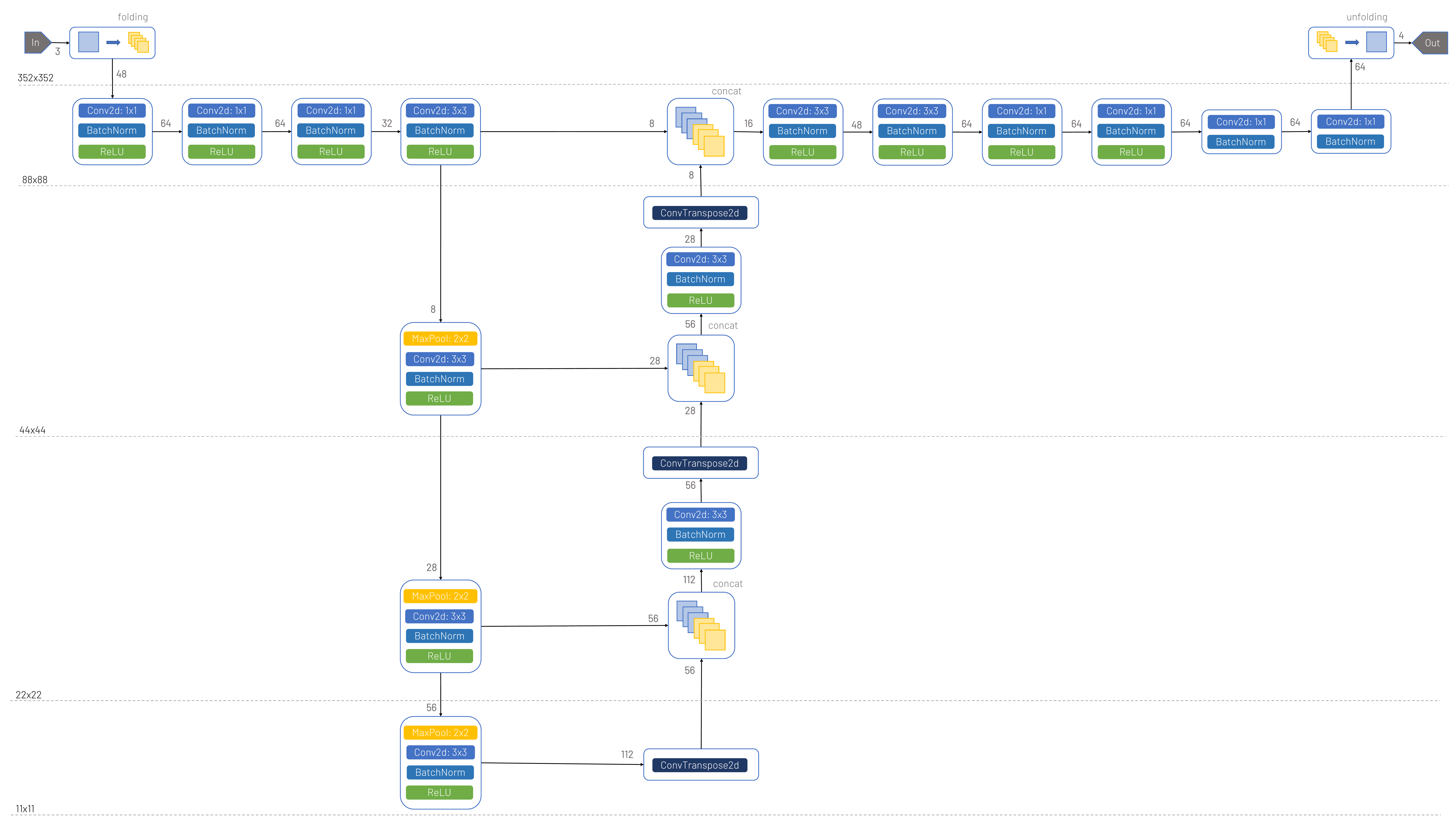}
  \caption{Proposed tiny U-net model architecture}
  \label{fig:unet_maxim}
\end{figure*}

Because of the data reshaping limitations of the chosen deployment platform, we utilize data folding only at the beginning layer of the proposed architecture. To reverse the folding action and create the segmentation map with the same resolution of the input image, data unfolding is applied at the end of the network. This operation is simply a tensor reshaping. In Fig. \ref{fig:unet_maxim}, the proposed UL-net architecture is illustrated. In the figure, the input image has three channels, and its size is 352x352. With data folding, a 48x88x88 shaped input tensor is obtained. In this spatial resolution, the architecture sequentially contains three Conv2d layers with 1x1 kernel and one Conv2d layer with 3x3 kernel. In the spatial resolution contraction path, there are three Conv2d 3x3 layers with a 2x2 MaxPool preceding each of them. After the contraction path, three ConvTranspose2d operations take place in the the expanding path. Between the contraction and expanding paths at each resolution, skip layers and channel-wise concatenations exist to facilitate information exchange and gradient propagation. After the resolution is increased back to 88x88, two Conv2d layers with 3x3 kernel and four Conv2d layers with 1x1 kernel are utilized to create the segmentation map.

All of the Conv2d layers contain batch normalization (BatchNorm), and before quantization-aware training (QAT) starts, the BatchNorm layers are fused into the weights and biases of the preceding Conv2d layers. The total number of parameters of the proposed architecture is 278,176 for four-class segmentation and 277,004 for two-class segmentation.

\section{Deployment Platform}
\label{sec:platform}
The proposed model is intended for deployment on CNN inference engines that work at the edge. As an example deployment platform, Analog Devices MAX78000 is chosen in this study. MAX78000 is an Ultra-Low-Power Arm Cortex-M4 processor with FPU-based MCU with CNN accelerator that targets battery-powered neural network-based applications at the edge \cite{max78000_datasheet}. MAX78000 has a weight memory of 442KB, and 1, 2, 4, or 8-bit weights are supported. As layers, it supports both Conv1D and Conv2D as convolution operations, 2D transposed convolution layer, linear layer, pooling layers, and element-wise operations. MAX78000 has 64 parallel processors. Each can only read from specific segments of the data memory but can write to the entire data memory. The CNN parallelization is performed at the input channels. Hence each input channel or group of input channels are processed by a different processor. The proposed data folding method leverages this feature by assigning the same load to a higher number of channels which results in an increase of parallelism.

\section{Experiments}
\label{sec:experiments}
\subsection{Dataset}
\label{subsec:dataset}
This study conducts experiments on two different datasets, and we present two possible edge applications. The first dataset, the Cambridge-driving Labeled Video Database (CamVid), provides ground truth labels that associate each pixel with one of 32 semantic classes \cite{CamVid}. Many researchers use this common dataset to understand complex street views, enabling different applications, especially for driving safety and navigation. We used three semantic classes for our application: `Building', `Sky', and `Tree', where the remaining classes fall into the class `Other'.
The CamVid dataset includes 601 960x720 color images, where 370 of them are used for training, and the rest are used for testing. Due to the memory limitations of the deployment platform, MAX78000, all images are cropped into intersecting 352x352 images. For the training set, the intersection ratio is kept around 40 percent to populate the training set sufficiently. For the test set, it is set to one percent to avoid biasing the accuracy results due to the duplicated pixels in the test set. Further, only the images with at least one of the selected three semantic classes are selected; the resulting set contains 4428 training samples and 1392 test samples. Finally, all images are folded with $\alpha$=4 as explained in Section \ref{sec:data_folding} to form images of size 48x88x88.

% There are 370 label image pairs in the training set and 233 pairs in the test set. With the data augmentation, 4428 training and 1392 test images are obtained. Dataset images have a resolution of (960x720) and are provided in RGB color format. In the training data augmentation, 12 images with size (352x352) are cropped from the original and label images. These square images are cropped starting from the top left corner and moving in an intersecting manner in both directions until original image borders are exceeded, like striding approach. The intersection among these square images is 150 and 168 pixels in x and y directions, respectively. In the test data augmentation, six images with size (352x352) are cropped similarly from the original and label images. The intersection areas, in this case are 54 and 0 pixels wide in x and y directions. All cropped square images are also folded into input images of size (48x88x88).

The second dataset, AISegment \cite{AISegment} is a human portrait segmentation dataset. The dataset includes 34,427 human portrait color images with a resolution of 600x800. The respective masks of the portrait images are provided in RGBA format. The alpha channel is either zero or one and is used for labeling each pixel with the `Background' or `Portrait' label. The dataset training-test split ratio is selected as $90\%-10\%$. The label distribution in both sets is around $43\%-57\%$ for the `Portrait' and `Background' classes, respectively. In the data augmentation phase, three overlapping images and corresponding matting images of size (600x600) are cropped from the original image, sliding in the y-direction. Each of the cropped images is then re-scaled to size (352x352). As a result of this cropping operation, the total number of images increases to 103,281. Finally, all images are folded with $\alpha$=4 to form images of size 48x88x88.

%These scaled images are then folded into input images of size (48x88x88).

%As demonstrated in Figure \ref{fig:folding_sample_img}, the input image contains 16 down-sampled versions of the input image.

% \begin{figure}[b]
%   \centering

%   \begin{subfigure}[t]{0.45\columnwidth}
%          \centering
%          \includegraphics[width=0.95\columnwidth]{figures/unetv7_folding_sample_image_high_res.png}
%          \caption{Sample image to be folded with size (3x352x352).}
%          \label{fig:folding_sample_high_res}
%      \end{subfigure}
%      \hfill
%      \begin{subfigure}[t]{0.45\columnwidth}
%          \centering
%          \includegraphics[width=0.9\columnwidth]{figures/unetv7_folding_sample_image_low_res.png}
%          \caption{First 3 channels of the folded image (48x88x88) refers to the down-sampled representation of the original image.}
%          \label{fig:folding_sample_low_res}
%      \end{subfigure}

%       \caption{Folding example on sample image.}
%          \label{fig:folding_sample_img}
% \end{figure}

\subsection{Training}
\label{subsec:training}

To be deployed on MAX78000, the model needs to be quantized to at most to 8-bit integers. Since simple post-quantization methods cause performance degradation \cite{max78000_nas_kws}, QAT is employed in this study. The QAT approach implements the fake (a.k.a. simulated) quantization approach of \cite{Jacob_2018_CVPR}. When training L\textsuperscript{3}U-net, an Adam optimizer is used with an initial learning rate of 0.001. For the CamVid dataset, the training takes 100 epochs, and QAT starts at epoch 40. The learning rate is halved two times at the 20\textsuperscript{th} and 80\textsuperscript{th} epochs. For the AISegment dataset, the training takes 200 epochs, QAT starts at epoch 150, and the learning rate is halved three times at the 50\textsuperscript{th}, 100\textsuperscript{th} and 140\textsuperscript{th} epochs. In this study, all weights are quantized to 8-bits, and batchnorm parameters are fused into convolutional layers before the QAT starts.

\subsection{Results}
\label{subsec:results}
The results of the proposed approach for the two selected datasets are summarized in Table \ref{tab:results}.

\newcolumntype{C}[1]{>{\centering\arraybackslash\hspace{0pt}}p{#1}}
\begin{table}
  \caption{Sample implementation results of proposed approach}
  \label{tab:results}
  \setlength{\tabcolsep}{2pt}
  \renewcommand{\arraystretch}{1.25}
  \begin{tabular}{| >{\centering}m{1.5cm} >{\centering}m{2cm} >{\centering}m{1cm} >{\centering}m{1.5cm} >{\centering\arraybackslash}m{1.5cm} |}
  %\begin{tabular}{| lcccC{0.1cm}C{0.1cm}C{0.1cm}C{0.1cm}C{0.1cm} |}
  \hline
  Dataset & Pixel-to-Pixel Accuracy (\%) & mIoU (\%) & Latency (ms) & Energy/Inf. (mJ) \\
  \hline
  \hline
  CamVid & 91.05 & 84.24 & 95.1 & 7.3 \\
  AISegment & 99.19 & 98.09 & 90.3 & 6.9 \\
  \hline
  \end{tabular}
\end{table}

\begin{figure}[h]
  \centering
  \includegraphics[width=1.0\linewidth]{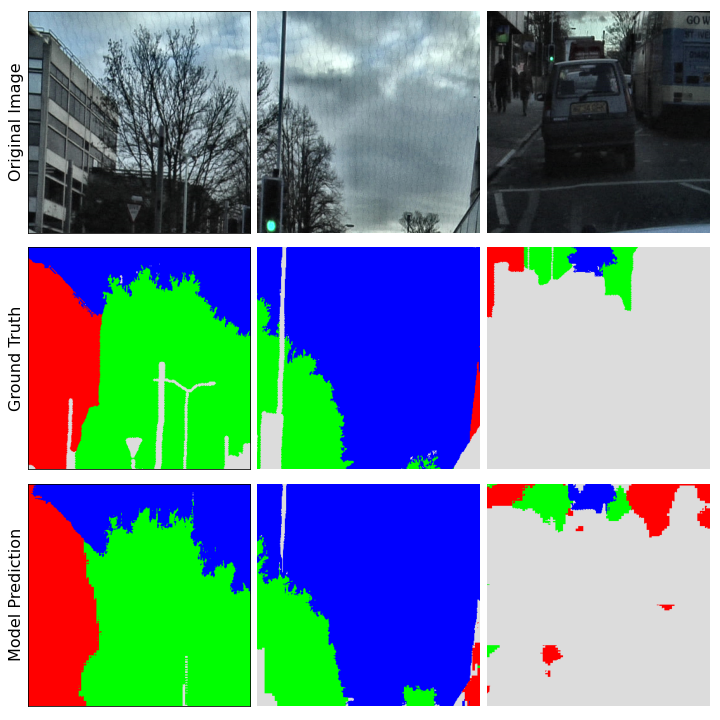}
  \caption{Sample model outputs for CamVid dataset.}
  \label{fig:CamVid_model_out}
\end{figure}

Fig. \ref{fig:CamVid_model_out} shows the results of L\textsuperscript{3}U-net for three sample images. As seen from these samples, the model accurately segments the regions, except for very tiny structures. For the four-class CamVid dataset, our approach provides better accuracy than many of the edge approaches in the literature AttendSeg \cite{wen2021attendseg} or EdgeSegNet \cite{edgesegnet}. As these works give their results for 32-class segmentation, it is not possible to compare the results in terms of accuracy, only model size and complexity of the operations. Our approach has 4x and 30x fewer parameters; 10x and 100x fewer MAC operations \cite{wen2021attendseg, edgesegnet}. In addition, with the proposed data folding approach, we were able to run the model on a battery-powered edge device, MAX78000, with an inference speed up to 10 fps.

\begin{figure}[h]
  \centering
  \includegraphics[width=0.9\linewidth]{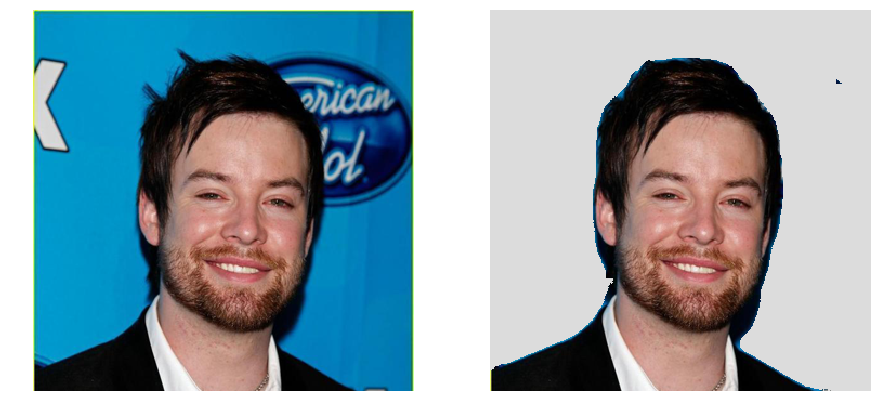}
  \caption{Sample model output for portrait segmentation.}
  \label{fig:aisegment_model_out}
\end{figure}

The model output for the AISegment dataset is the binary class label of each pixel, and Fig. \ref{fig:aisegment_model_out} includes a model input image and removed background by L\textsuperscript{3}U-net model. A reference model, \cite{ref_study_portrait_seg}, provides 98\% mIoU for 224x224 image with a floating-point model that has 8x more parameters than ours. In that sense, L\textsuperscript{3}U-net is a more effective approach requiring less memory and less computation for larger input data.

% For AISegment, our approach provides similar accuracy than the one in the reference study \cite{ref_study_portrait_seg} while our model has 8x less 8-bit integer parameters. The input resolution of the reference model is also 224x224 which

% As listed in Table \ref{tab:results}, model developed using the AISegment dataset achieves 99.19\% accuracy and 98.09\% mIoU. A reference study on portrait segmentation using the same AISegment dataset \cite{ref_study_portrait_seg} achieves 97\% accuracy and 98\% mIoU with input images of resolution (224x224) using a Mobile-Unet variant architecture.

% We also measured the latency and the energy consumption on device for both models and results are presented in Table \ref{tab:results}. As the only difference among the models is in the output layer resulting from the difference in the number of labels, these performance metrics are measured very close values.

%model output mask image and the mask image overlayed into the input image with alpha value 0.2.

Lastly, we provide energy measurements of the L\textsuperscript{3}U-net model on MAX78000. According to the results, with an AA alkaline battery, it is possible to calculate over 1.5 million inferences using the proposed model and setup. To the best of our knowledge, no other study gives these values for semantic segmentation at the edge; so we provide these results for future studies.

\section{Conclusion}
\label{sec:conclusion}
We presented \ournet, a lightweight U-net-based fully CNN model which is also deployed to a tiny edge device, MAX78000. We also proposed a data folding approach that reduces the latency and enables processing of higher resolution input images for a multi-core CNN inference engine, MAX78000. The experiments performed show that \ournet is capable of providing 91\% and 99\% accuracy for four- and two-class segmentation examples at a speed of 10 fps.

\begin{acks}
The authors would like to thank Robert Muchsel, Brian Rush, and other members of the AI Development Group at Analog Devices that contributed to this work.
\end{acks}

\bibliographystyle{ACM-Reference-Format}
\bibliography{acmart}

\end{document}